\documentclass[10pt,twocolumn,letterpaper]{article}

\usepackage[final]{iccv
}          

\usepackage{algorithm} 
\usepackage{algpseudocode} 
\usepackage{subcaption}
\usepackage{amsmath} 
\usepackage{graphicx} 
\usepackage{lipsum} 
\usepackage{colortbl}
\usepackage{multirow}

\definecolor{iccvblue}{rgb}{0.21,0.49,0.74}
\usepackage[pagebackref,breaklinks,colorlinks,allcolors=iccvblue]{hyperref}

\def\paperID{9146} 
\def\confName{ICCV}
\def\confYear{2025}

\title{RainFusion: Adaptive Video Generation Acceleration via Multi-Dimensional Visual Redundancy}

\author{
    Aiyue Chen\textsuperscript{1*}, Bin Dong\textsuperscript{1*}, Jingru Li\textsuperscript{1} \\
    Jing Lin\textsuperscript{1} , Kun Tian\textsuperscript{1} , Yiwu Yao\textsuperscript{1}, Gongyi Wang\textsuperscript{1} \\[4pt]
    \textsuperscript{1}Huawei Technologies Co., Ltd
}

\begin{document}
\maketitle
\renewcommand{\thefootnote}{\fnsymbol{footnote}} 
\footnotetext[1]{These authors contributed equally.}
\begin{abstract}
Video generation using diffusion models is highly computationally intensive, with 3D attention in Diffusion Transformer (DiT) models accounting for over 80\% of the total computational resources. In this work, we introduce {\bf RainFusion}, a novel training-free sparse attention method that exploits inherent sparsity nature in visual data to accelerate attention computation while preserving video quality. Specifically, we identify three unique sparse patterns in video generation attention calculations--Spatial Pattern, Temporal Pattern and Textural Pattern. The sparse pattern for each attention head is determined online with negligible overhead (\textasciitilde\,0.2\%) with our proposed {\bf ARM} (Adaptive Recognition Module) during inference. Our proposed {\bf RainFusion} is a plug-and-play method, that can be seamlessly integrated into state-of-the-art 3D-attention video generation models without additional training or calibration. We evaluate our method on leading open-sourced models including HunyuanVideo, OpenSoraPlan-1.2 and CogVideoX-5B, demonstrating its broad applicability and effectiveness. Experimental results show that RainFusion achieves over {\bf 2\(\times\)} speedup in attention computation while maintaining video quality, with only a minimal impact on VBench scores (-0.2\%).
\end{abstract}    
\section{Introduction}
\label{sec:intro}

Diffusion models have become the leading approach in video generation, demonstrating exceptional performance and broad applicability \cite{blattmann2023stablevideodiffusionscaling} \cite{henschel2024streamingt2vconsistentdynamicextendable} \cite{polyak2024moviegencastmedia} \cite{OpenAISora2024}. Initially built on U-Net architectures \cite{blattmann2023stablevideodiffusionscaling} \cite{henschel2024streamingt2vconsistentdynamicextendable}, the field has transitioned to Diffusion Transformers (DiTs), which now serve as the mainstream approach owing to their enhanced performance and scalability. This architectural evolution has further advanced with the adoption of 3D full-sequence attention mechanisms \cite{OpenAISora2024} \cite{polyak2024moviegencastmedia}, replacing the previously dominant 2D+1D spatial-temporal attention (STDiT) \cite{ma2024lattelatentdiffusiontransformer} that separately computes spatial and temporal attention alternatively. Although these advancements have enhanced modeling capabilities, they also impose substantial computational challenges, particularly in attention computation.

\begin{figure}[t]
  \centering
  \includegraphics[width=1.0\linewidth]{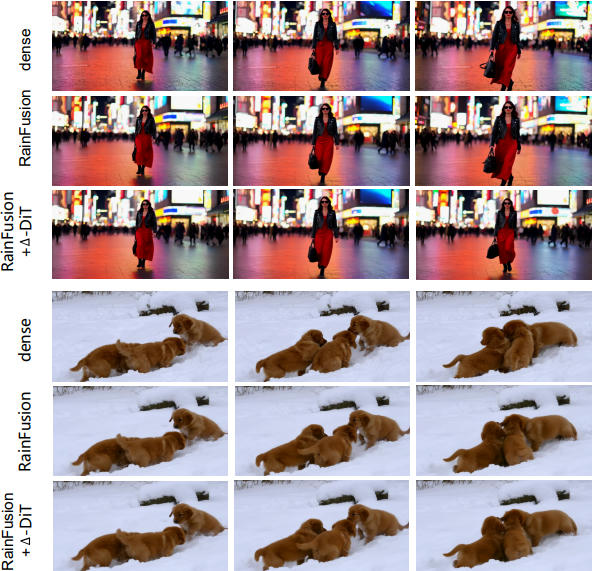}

   \caption{ HunyuanVideo 720p RainFusion results. RainFusion and RainFusion combined with $\Delta$-DiT shows good visual quality and high similarity to dense results. Upper prompt: ``A stylish woman walks down a Tokyo street filled with warm glowing neon and animated city signage.". Lower prompt: ``A litter of golden retriever puppies playing in the snow. Their heads pop out of the snow, covered in.".}
   \label{fig:hy_720p}
\end{figure}

\begin{figure*}[t]
    \centering
    \includegraphics[width=1.0\linewidth]{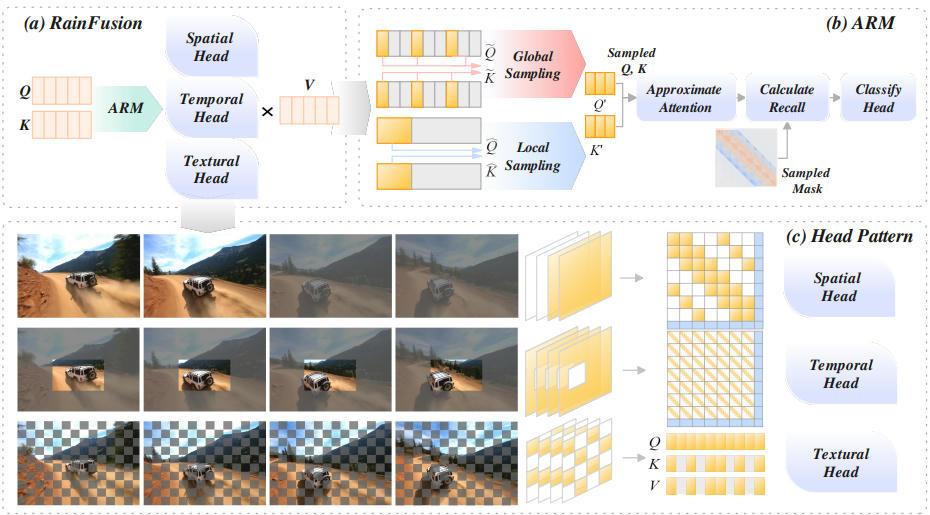}
    \caption{(a) RainFusion pipeline including Adaptive Recognition Module(ARM) and applying sparse pattern to Flash Attention. (b) ARM determine the pattern using subset of query and key to calculate approximates attention score and applying the predefined pattern mask to get attention recall to determine the head category. The sampled queries $Q'$ and keys $K'$ are either sourced from the tokens of the first frame or obtained by sampling from the full set of tokens with equal intervals. 
    (c) The three head sparse pattern. These three heads respectively concentrate on portraying global spatial details with local temporal information, local spatial details with global temporal information, and high-level textural information.}
    \label{fig:RainFusion}
\end{figure*}

The computational complexity of these models scales quadratically with the sequence length, expressed as $O(s^2 t^2)$, where $s$ and $t$ represent the spatial and temporal dimensions, respectively. This scaling poses a substantial bottleneck, as evidenced by the deployment of Open-Sora-Plan 1.2 \cite{OpenSoraPlan2024} on a single A100 GPU, which requires approximately 48 minutes to generate a 4-second 720p video. Profiling analysis demonstrates that the attention mechanism consumes over 80\% of total computation, making it the principal performance bottleneck in the video generation pipeline.

To improve the computational efficiency of video generation models, researchers have developed two key algorithms: (1) sampling optimization techniques that reduce the number of required inference steps through adaptive sampling schedules \cite{li2023autodiffusiontrainingfreeoptimizationtime} \cite{sabour2024alignstepsoptimizingsampling}, and (2) caching mechanisms that exploit redundancy by reusing features across adjacent timesteps \cite{chen2024deltadittrainingfreeaccelerationmethod} \cite{kahatapitiya2024adaptivecachingfastervideo} \cite{liu2024timestepembeddingtellsits}. Sampling optimization techniques are inherently limited by their dependency on post-training adjustments, limiting their practical applicability. Furthermore, both sampling optimization and caching algorithms necessitate models to operate with a relatively large number of inference steps, as their effectiveness relies heavily on sufficient redundancy between consecutive timesteps.
Despite these advancements, optimizing the attention mechanism has not yet been explored in depth. DiTFastAttn\cite{yuan2024ditfastattnattentioncompressiondiffusion} use brute-force sliding-window mask and use residual cache to compensate quality loss. SVG\cite{xi2025sparse} ignores model generality and the inherent visual feature in video. 

\begin{figure}[t]
  \centering
  \includegraphics[width=1.0\linewidth]{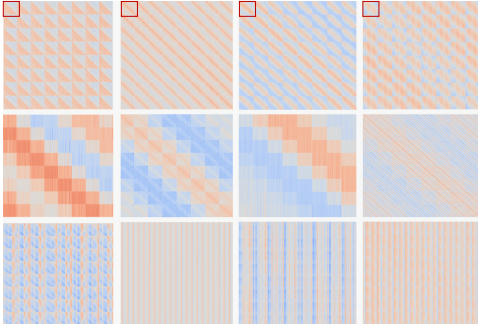}

   \caption{The attention sparsity pattern with the vertical axis and the horizontal axis representing query and key respectively. The first row depicts the temporal sparsity pattern, which models the same spatial location across different frames (with the red box in the upper-left corner highlighting the basic repeated pattern). The second row shows the spatial sparsity pattern, focusing on all locations in neighboring frames. The third row presents a conventional full-attention head, for which we propose a sophisticated textural sparse attention mechanism.}
   \label{fig:sparse_pattern}
\end{figure}

In this work, we introduce a novel sparse attention mechanism that effectively leverages two key characteristics of video generation: (1) the inherent spatial-temporal redundancy in video, and (2) the importance of specific image texture. We observe that there exists three types of sparse pattern in attention, one for temporal pattern which attends to the same spatial location in different frames, one for spatial pattern which models all spatial location in consecutive frames, the other for detailed texture of video frames. As shown in Fig.\ref{fig:sparse_pattern}, it is the attention score-map of some heads with the vertical axis and the horizontal axis representing q and k respectively. The first row captures local repetitive patterns within each window, which we define as the {\bf Temporal Head}, indicating that certain heads consistently attend to the same locations across different frames. The second row reveals more global patterns across neighboring frames, which we term the {\bf Spatial Head}. The third row highlights {\bf Textural Heads}, where important tokens are attended to by all query tokens. We determine the sparse pattern for each head online using Adaptive Recognition Module(ARM) which only introduce $\frac{1}{t^2}$ overhead where t represents frame number in latent space. The overall pipeline is shown in Fig.\ref{fig:RainFusion}. We first determine sparse pattern of different head online using global or local sampling, and then calculate attention using their respective sparse pattern. 

Extensive experiments on different video generation models including OpenSoraPlan-1.2 \cite{OpenSoraPlan2024}, HunyuanVideo-13B \cite{kong2025hunyuanvideosystematicframeworklarge}, CogVideoX-5B \cite{yang2024cogvideoxtexttovideodiffusionmodels} prove the generality and effectiveness of RainFusion. The contributions of this paper include:

\noindent 
\begin{itemize}
    \item We present {\bf RainFusion}, a novel plug-and-play framework that leverages tri-dimensional sparsity across spatial, temporal, and textural domains to optimize video diffusion models. The proposed method dynamically determines sparse patterns through online estimation, effectively exploiting the intrinsic redundancy inherent in video data. The name {\bf RainFusion} is derived from the observation that the sparse patterns resemble the continuous, interconnected lines formed by rain.
    \item We put forward a simple but potent sparse pattern estimation method {\bf ARM} that entails minimal computational cost (\textasciitilde\,0.2\% overhead), thereby rendering our RainFusion highly efficient.
    \item RainFusion can be applied to many SOTA video generation models, OpenSoraPlan-1.2 \cite{OpenSoraPlan2024}, HunyuanVideo-13B \cite{kong2025hunyuanvideosystematicframeworklarge}, CogVideoX-5B \cite{yang2024cogvideoxtexttovideodiffusionmodels} with over 2x speedup in attention at negligible quality loss (-0.2\% VBench score) as shown in Fig. \ref{fig:cmp_with_sota}.
\end{itemize}



\medskip


\section{Related Work}
\label{sec:formatting}

\subsection{Diffusion Models}
Diffusion models \cite{ho2020denoising,nichol2021improved,peebles2023scalable,dhariwal2022guided,chen2024pixart,zheng2024open,ma2025inference} have surpassed Generative Adversarial Networks (GANs) in generative tasks by iteratively reversing a noisy process to synthesize data, such as images, through progressive denoising. These models typically use U-Net \cite{rombach2022high,podell2023sdxl,blattmann2023stable} or transformer-based architectures \cite{peebles2023scalable}, with the latter gaining prominence in vision applications, as seen in DiT (Diffusion Transformers) \cite{peebles2023scalable} for data distribution modeling and PixArt-$\Sigma$\cite{chen2024pixart} for 4K image generation. Furthermore, diffusion models have been extended to video synthesis \cite{blattmann2023stable}, with two main approaches emerging: (1) the 2D+1D STDiT structure, used in OpenSora \cite{zheng2024open}, and (2) the 3D full-sequence attention mechanism, employed by Sora \cite{OpenAISora2024}, Open-Sora-Plan 1.2 \cite{OpenSoraPlan2024}, CogVideoX \cite{yang2024cogvideoxtexttovideodiffusionmodels}, and Hunyuan Video \cite{kong2025hunyuanvideosystematicframeworklarge}. These developments underscore the versatility and scalability of diffusion models in tackling increasingly complex generative tasks.

\subsection{Sparse Attention in Transformers}


In Transformer-based large models, the quadratic complexity of matrix multiplication $QK^T$ in attention mechanisms drives high computational costs. To address this, recent research exploits sparsity in attention maps \cite{jiang2024minference10acceleratingprefilling,tang2024razorattention,zhang2024oats}, and some research use techniques like token pruning \cite{smith2024todo,tian2024u,wang2024attention} and token merging \cite{bolya2023token,wu2024importance,tran2025accelerating} to reduce sequence length and improve inference efficiency. Some methods employ dynamic sparse attention \cite{jiang2024minference10acceleratingprefilling} or merge sparse tokens \cite{tang2024razorattention} to accelerate LLM inference. Similarly, in vision-specific models like ViTs and DiTs, sparsity is leveraged through dynamic activation pruning \cite{chen2023sparsevit}, pixel downsampling \cite{smith2024todo}, and KV matrix downsampling \cite{tian2024u}, while video generation adapts token merging via bipartite soft matching \cite{bolya2023token}, importance sampling \cite{wu2024importance}, and spectrum-preserving techniques \cite{tran2025accelerating}. These advancements highlight the broad potential of sparse attention to enhance efficiency across diverse domains.

\subsection{Attention Sharing and Cache}


When accelerating diffusion model inference, cache methods leverage attention map similarity between adjacent denoising timesteps \cite{liu2024timestepembeddingtellsits,wimbauer2024cache,ma2024learning}. For example, $\Delta$-DiT \cite{chen2024delta} introduces a tailored caching method for DiT acceleration, while DeepCache \cite{ma2023deepcache} and TGATE \cite{liu2024faster} reduce redundant calculations by layer-wise attention similarities. Recent methods further optimize performance by caching model outputs \cite{liu2024timestepembeddingtellsits} or dynamically adjusting caching strategies \cite{kahatapitiya2024adaptivecachingfastervideo}. Additionally, techniques like DiTFastAttn \cite{yuan2024ditfastattnattentioncompressiondiffusion} combine sparse attention with caching, exploiting spatial, temporal, and conditional redundancies for efficient attention compression. These advancements demonstrate the potential of integrating sparse attention and caching to enhance the scalability and speed of diffusion model inference.

\textbf{Recent Work.} SVG \cite{xi2025sparse} advances sparse attention research by analyzing spatial and temporal attention sparsity in DiTs and proposing a training-free online profiling strategy. However, they classify attention heads only into temporal and spatial groups, neglecting irregular attention patterns in video generation. Unlike SVG, our work focuses on irregular attention heads to capture fine-grained textural details for improved video generation.

\section{Methodology}
\label{sec:method}
In this section, we introduce \textbf{RainFusion}, a training-free adaptive algorithm, designed to exploit the computing sparsity in 3D full attention to accelerate video generation.

\subsection{Preliminary}

Existing video generation models utilize 3D full attention mechanisms, which jointly capture both spatial and temporal dependencies to elevate generation quality. However, it comes at a high computational cost.

We define the shape of the latent video as $(H,W,T)$. In 3D full attention, the video sequence is formed by flattening T sub-sequences, each sub-sequence represents a single frame of length 
$H \times W$.
We denote \((Q,K,V)\in \mathbb{R}^{N \times d}\) as the query, key, and value tokens,  respectively, and define 
M as the attention mask with shape \(N \times N\), where \(N=H\times W\times T\) and \(d\) is the hidden dimension of each head.
The bidirectional 3D full attention can be formulated as follows:
\begin{equation}
S(Q,K,M) \gets Softmax(\frac{QK^{T}}{\sqrt{d}} + M )
\label{equa:softmax}
\end{equation}
\begin{equation}
Attn(Q,K,V,M) \gets S(Q,K,M)V
\end{equation}

The computation complexity is \(O(N^{2})\).
While 3D full attention mechanisms are inherently dense, our analysis reveals discernible computational sparsity patterns across attention heads. As shown in Fig.\ref{fig:sparse_pattern} and Fig. \ref{fig:RainFusion} (c), we classify these specialized heads into three categories: Spatial head, Temporal head, and Textural head.

\subsection{Attention Head Mechanism Design}
\label{head classification}
\paragraph{Spatial Head}
The Spatial Head exhibits global spatial dependencies within individual frames while capturing localized temporal dependencies across the full sequence.   
This characteristic indicates that the Spatial Head emphasizes both the completeness of individual frames and the overall coherence among adjacent or key frames. 
Consequently, it suggests that certain non-key frames hold relatively less significance and can be excluded from the attention calculation.
\begin{equation}
Attn_{spatial}\gets Attn(Q_{f},K_{\{f^\prime\}},V_{\{f^\prime\}},M_{spatial})
\end{equation}
Here, \(\{f^\prime\}\) denotes the set of significant frames for the \(f^{th}\) frame. Therefore, a global striped  attention mask \(M_{spatial}\) is designed as depicted in Fig.\ref{fig:RainFusion} (c). A continuous sub-sequence of a frame is defined as a window segment. The positions of these window segments determine both the key frame attended by the attention mechanism and the resulting computational gains.


\paragraph{Temporal Head}
Contrary to the Spatial Head, the Temporal Head demonstrates locality within a single-frame sub-sequence in spatial domain, while exhibits a global characteristic in whole temporal domain.
The Temporal Head is particularly attentive to the correlation between the same local regions across different video frames. Its primary focus is on creating regional details that maintain spatial continuity. This unique property can lead to the manifestation of local sparsity within a single-frame sub-sequence and periodic sparsity throughout the entire sequence, as shown in Fig.\ref{fig:RainFusion} (c).
\begin{figure}[t]
  \centering
  \includegraphics[width=1.0\linewidth]{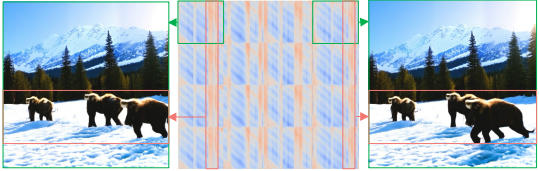}

   \caption{The above figure shows the attention score map of a typical textural head. The green region represents a single frame. Notably, the pink region, characterized by high attention score for most \(Q\), coincides with the motion regions emphasized by the prompt.}
   \label{fig:textural}
\end{figure}
\paragraph{Textural Head}
It becomes evident that certain content holds significant importance throughout the entire video, particularly those parts intricately linked to high-level textural description, which is shown in Fig.\ref{fig:textural}. This is manifested in the fact that some specific \(K, V\) consistently receive high attention scores for most \(Q\). As a result, while the distribution of tokens is sparse, it is challenging to identify a regular attention mask that can effectively adapt to this sparsity state.
Based on the above considerations, we condense the \(K, V\) sequence approximately by referring to the property of image downsampling. The \(K, V\) sequence will be rearranged and tokens will be retained in a checkerboard - interleaving pattern in spatial domain, as depicted in Fig.\ref{fig:RainFusion} (c).
\begin{multline}
C = \{a_{ij} \mid ((i \bmod \tau = k) \land (j \bmod \tau = k)), \\
    0 \leq i < H, 1 \leq j < W, 0 \leq k < \tau \}
\end{multline}
\begin{equation}
Attn_{Irregular}\to Attn(Q,K_{\{C\}},V_{\{C\}},M_{init})
\end{equation}
\(C\) indicates the set of chosen \(K, V\) token indexes, \(\tau\) represents the stride of the checkerboard, and \(M_{\text{init}}\) be the all-zero mask. The checkerboard format ensures that information from each discarded token in the spatial domain can be implicitly generated by referring to the four nearest remaining tokens. Additionally, we opt to directly retain or discard some tokens rather than averaging them. This is because averaging would obscure the intrinsic information of tokens, making it challenging to implicitly supply the correct information for discarded tokens.

\subsection{Adaptive Recognition Module(ARM)}
\label{method:arm}
As described in Sec 3.2, RainFusion categorizes all heads into three distinct types: Spatial Head, Temporal Head, and Textural Head. However, we find that the pattern of each head is highly dynamic. For instance, factors such as input prompts and sampling steps all influence the characteristics of each head. Given these considerations, we introduce an Adaptive Recognition Module (ARM). This module is designed to do online and adaptive classification for all heads with minimal computational cost.

We first acquire the approximate attention score, and then compute the masked attention recall.
As illustrated in Fig \ref{fig:RainFusion} (b), in local sampling, we select the tokens of the first frame sub-sequence as \(\widehat{Q}, \widehat{K}\).In the case of global sampling, we sample tokens at equal intervals \(\omega\) as \(\widetilde{Q}, \widetilde{K}\). We utilize the downsampled sequences to calculate the attention score, which serves as an approximation of the overall attention score. Then we compute the masked recall based on the approximate score:
\begin{equation}
R^\prime \gets Recall(Q^\prime, K^\prime, M^\prime) = \frac{S(Q^\prime, K^\prime, M^\prime)}{S(Q^\prime, K^\prime, M_{init})}
\end{equation}

\(Q^\prime, K^\prime\) represent the downsampled sequences.\(M^\prime\) denotes the attention mask derived by downsampling either \(M_{spatial}\) or \(M_{temporal}\) in accordance with the  corresponding token downsampling rules. \(S\) means softmax operation as shown in Equation
Attention recall means the proportion of valid information that can be preserved under the current pattern mask. Through this method, we are able to adaptively and efficiently determine the category of each head online with minimal computational overhead.

Algorithm 1 provides a detailed introduction to the process of the Adaptive Recognition Module (ARM).

\section{Experiments}

\begin{table*}
  \centering 
  \resizebox{\textwidth} {!}{ 
      \begin{tabular}{llccccc}
      \toprule
                                                                  &                                                            &                                        & {\color[HTML]{333333} }                                & {\color[HTML]{333333} }                                 & {\color[HTML]{333333} }                              & {\color[HTML]{333333} }                          \\
        \multirow{-2}{*}{Model}                                   & \multirow{-2}{*}{Method}                                   & \multirow{-2}{*}{Loss}                 & \multirow{-2}{*}{{\color[HTML]{333333} Quality Score$\uparrow$}} & \multirow{-2}{*}{{\color[HTML]{333333} Semantic Score$\uparrow$}} & \multirow{-2}{*}{{\color[HTML]{333333} Total Score$\uparrow$}} & \multirow{-2}{*}{{\color[HTML]{333333} Speedup}} \\
        \midrule
        {\color[HTML]{333333} }                                   & {\color[HTML]{000000} baseline}                            & /                                      & 82.04                                                  & 70.7                                                    & 79.77                                                & 1.0\(\times\)                                             \\
        {\color[HTML]{333333} }                                   & {\color[HTML]{000000} $\Delta$-DiT}                               & -5.37                                  & 76.56                                                  & 65.76                                                   & 74.4                                                 & 1.81\(\times\)                                            \\
        {\color[HTML]{333333} }                                   & {\color[HTML]{000000} DiTFastAttn}                         & -5.36                                  & 77.94                                                  & 60.27                                                   & 74.41                                                & 1.52\(\times\)                                            \\
        \multirow{-4}{*}{{\color[HTML]{333333} CogvideoX-5B}}     & \cellcolor[HTML]{ECFAE8}{\color[HTML]{000000} RainFusion}  & \cellcolor[HTML]{ECFAE8}\textbf{-0.28} & \cellcolor[HTML]{ECFAE8}\textbf{81.64}                 & \cellcolor[HTML]{ECFAE8}\textbf{70.91}                  & \cellcolor[HTML]{ECFAE8}\textbf{79.49}               & \cellcolor[HTML]{ECFAE8}\textbf{1.85\(\times\)}           \\
        \midrule
        {\color[HTML]{333333} }                                   & {\color[HTML]{000000} baseline}                            & /                                      & 79.6                                                   & 38.01                                                   & 71.28                                                & 1.0\(\times\)                                             \\
        {\color[HTML]{333333} }                                   & {\color[HTML]{000000} $\Delta$-DiT}                               & -0.93                                  & 78.51                                                  & 37.7                                                    & 70.35                                                & 1.81\(\times\)                                            \\
        {\color[HTML]{333333} }                                   & {\color[HTML]{000000} DiTFastAttn}                         & -2.31                                  & 77.56                                                  & 34.59                                                   & 68.97                                                & 1.42\(\times\)                                            \\
        \multirow{-4}{*}{{\color[HTML]{333333} OpenSoraPlan-1.2}} & \cellcolor[HTML]{ECFAE8}{\color[HTML]{000000} RainFusion}  & \cellcolor[HTML]{ECFAE8}\textbf{-0.32} & \cellcolor[HTML]{ECFAE8}\textbf{79.08}                 & \cellcolor[HTML]{ECFAE8}\textbf{38.44}                  & \cellcolor[HTML]{ECFAE8}\textbf{70.95}               & \cellcolor[HTML]{ECFAE8}\textbf{1.91\(\times\)}           \\
        \midrule
        {\color[HTML]{333333} }                                   & {\color[HTML]{000000} baseline}                            & /                                      & 84.11                                                  & 70.12                                                   & 81.31                                                & 1.0\(\times\)                                             \\
        {\color[HTML]{333333} }                                   & {\color[HTML]{000000} $\Delta$-DiT}                               & -0.87                                  & 83.27                                                  & 69.08                                                   & 80.44                                                & 1.81\(\times\)                                            \\
        {\color[HTML]{333333} }                                   & {\color[HTML]{000000} DiTFastAttn}                         & -1.14                                  & 82.87                                                  & 69.37                                                   & 80.17                                                & 1.78\(\times\)                                            \\
        {\color[HTML]{333333} }                                   & \cellcolor[HTML]{ECFAE8}{\color[HTML]{333333} RainFusion}  & \cellcolor[HTML]{ECFAE8}-0.4           & \cellcolor[HTML]{ECFAE8}83.77                          & \cellcolor[HTML]{ECFAE8}69.46                           & \cellcolor[HTML]{ECFAE8}80.91                        & \cellcolor[HTML]{ECFAE8}1.89\(\times\)                    \\
        {\color[HTML]{333333} }                                   & \cellcolor[HTML]{ECFAE8}{\color[HTML]{333333} RainFusion+} & \cellcolor[HTML]{ECFAE8}\textbf{-0.19} & \cellcolor[HTML]{ECFAE8}\textbf{83.79}                 & \cellcolor[HTML]{ECFAE8}\textbf{70.46}                  & \cellcolor[HTML]{ECFAE8}\textbf{81.12}               & \cellcolor[HTML]{ECFAE8}\textbf{1.84\(\times\)}        \\
        \multirow{-6}{*}{{\color[HTML]{333333} HunyuanVideo}}     & \cellcolor[HTML]{ECFAE8}RainFusion+ \& $\Delta$-DiT               & \cellcolor[HTML]{ECFAE8}-0.49          & \cellcolor[HTML]{ECFAE8}83.43                          & \cellcolor[HTML]{ECFAE8}70.35                           & \cellcolor[HTML]{ECFAE8}80.82                        & \cellcolor[HTML]{ECFAE8}2.37\(\times\)                \\
        \bottomrule
        \end{tabular}
  }
  \caption{Comparison with state-of-the-art algorithms. }
  \label{tab:cmp_with_sota}
\end{table*}


\begin{table*}
  \centering
    \resizebox{1.0\textwidth} {!}{
        \begin{tabular}{lccccccccccc}
        \toprule
                                           &                           &                           &                           &                           &                                                                           &                                                                                 &                                                                                    &                                                                               &                                                                            &                                                                               &                                                                             \\
        \multirow{-2}{*}{Model}            & \multirow{-2}{*}{S}       & \multirow{-2}{*}{T}       & \multirow{-2}{*}{Te}      & \multirow{-2}{*}{L}       & \multirow{-2}{*}{\begin{tabular}[c]{@{}c@{}}Average \\ Loss\end{tabular}} & \multirow{-2}{*}{\begin{tabular}[c]{@{}c@{}}Subject\\ Consistency$\uparrow$\end{tabular}} & \multirow{-2}{*}{\begin{tabular}[c]{@{}c@{}}Background\\ Consistency$\uparrow$\end{tabular}} & \multirow{-2}{*}{\begin{tabular}[c]{@{}c@{}}Motion\\ Smoothness$\uparrow$\end{tabular}} & \multirow{-2}{*}{\begin{tabular}[c]{@{}c@{}}Dynamic\\ Degree$\uparrow$\end{tabular}} & \multirow{-2}{*}{\begin{tabular}[c]{@{}c@{}}Aesthetic\\ Quality$\uparrow$\end{tabular}} & \multirow{-2}{*}{\begin{tabular}[c]{@{}c@{}}Imaging\\ Quality$\uparrow$\end{tabular}} \\
        \midrule
                                           &                           &                           &                           &                           & /                                                                         & 93.48                                                                           & 95.24                                                                              & 97.19                                                                         & 45.83                                                                      & 58.12                                                                         & 64.79                                                                       \\
                                           & \checkmark & \checkmark                         &                           & \checkmark                         & -1.05                                                                     & 92.87                                                                           & 94.65                                                                              & 97.47                                                                         & 45.83                                                                      & 56.58                                                                         & 60.91                                                                       \\
                                           & \cellcolor[HTML]{ECFAE8}\checkmark & \cellcolor[HTML]{ECFAE8}\checkmark & \cellcolor[HTML]{ECFAE8}\checkmark & \cellcolor[HTML]{ECFAE8}\checkmark & \cellcolor[HTML]{ECFAE8}\textbf{-0.18}                                    & \cellcolor[HTML]{ECFAE8}\textbf{93.27}                                          & \cellcolor[HTML]{ECFAE8}95.31                                                      & \cellcolor[HTML]{ECFAE8}97.23                                                 & \cellcolor[HTML]{ECFAE8}\textbf{45.83}                                     & \cellcolor[HTML]{ECFAE8}\textbf{58.11}                                        & \cellcolor[HTML]{ECFAE8}\textbf{63.80}                                      \\
        \multirow{-4}{*}{CogvideoX-5B}     & \checkmark                         & \checkmark                         & \checkmark                         &                           & -0.42                                                                     & 93.08                                                                           & \textbf{95.40}                                                                     & \textbf{97.27}                                                                & 45.83                                                                      & 57.17                                                                         & 63.36                                                                       \\
        \midrule
                                           &                           &                           &                           &                           & /                                                                         & 94.65                                                                           & 95.19                                                                              & 99.40                                                                         & 41.67                                                                      & 56.84                                                                         & 57.67                                                                       \\
                                           & \checkmark                         & \checkmark                         &                           & \checkmark                         & -1.29                                                                     & 92.53                                                                           & \textbf{94.72}                                                                     & 98.94                                                                         & 43.75                                                                      & 55.05                                                                         & 52.66                                                                       \\
                                           & \cellcolor[HTML]{ECFAE8}\checkmark & \cellcolor[HTML]{ECFAE8}\checkmark & \cellcolor[HTML]{ECFAE8}\checkmark & \cellcolor[HTML]{ECFAE8}\checkmark & \cellcolor[HTML]{ECFAE8}\textbf{1.03}                                     & \cellcolor[HTML]{ECFAE8}\textbf{93.22}                                          & \cellcolor[HTML]{ECFAE8}94.31                                                      & \cellcolor[HTML]{ECFAE8}99.15                                                 & \cellcolor[HTML]{ECFAE8}\textbf{52.08}                                     & \cellcolor[HTML]{ECFAE8}\textbf{55.67}                                        & \cellcolor[HTML]{ECFAE8}\textbf{57.17}                                      \\
        \multirow{-4}{*}{OpenSoraPlan-1.2} & \checkmark                         & \checkmark                         & \checkmark                         &                           & 0.33                                                                      & 92.73                                                                           & 94.69                                                                              & \textbf{99.18}                                                                & 50.00                                                                      & 55.12                                                                         & 55.71  \\                           
        \bottomrule                                         
        \end{tabular}
    }
  \caption{Ablation Results. RainFusion with three pattern and local estimation achieves the best result. We denote S, T, Te, L as spatial head, temporal head, textural head and use local sampling in estimating local pattern recall, respectively.}
  \label{tab:ablation}
\end{table*}

\begin{figure*}[!]
  \centering
  \includegraphics[width=1.0\linewidth]{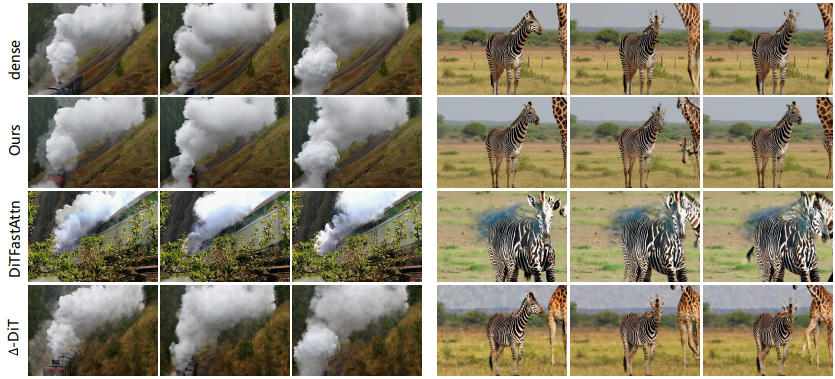}
   \caption{Video Comparison using CogVideoX-5B with different accelerating algorithms. Left prompt: ``A steam train moving on a mountainside." Right Prompt: ``a zebra on the left of a giraffe, front view.".}
   \label{fig:cmp_with_sota}
\end{figure*}

\subsection{Settings}
\paragraph{Models}
We evaluate RainFusion on three widely adopted video generation models: OpenSoraPlan-1.2 \cite{OpenSoraPlan2024}, HunyuanVideo-13B \cite{kong2025hunyuanvideosystematicframeworklarge} and CogVideoX-5B \cite{yang2024cogvideoxtexttovideodiffusionmodels}. For HunyuanVideo and OpenSoraPlan-1.2, we generate 125 and 93 frames at 480p resolution, with latent dimensions of 

\noindent 
\begin{minipage}{\columnwidth} 
    \begin{algorithm}[H] 
        \caption{Adaptive Recognition Module(ARM)} 
        \begin{algorithmic}
            \State \textbf{Input} \(Q,K,M_{spatial},M_{temporal}\)
            \State \textbf{Output} \(H\)  \#\ \textit{Head Category}
            \State \(\widehat{Q}, \widehat{K}, \widehat{M}_{temporal} \gets Local Sampling(Q,K,M_{temporal})\)
            \State \(\widetilde{Q}, \widetilde{K}, \widetilde{M}_{spatial} \gets Global Sampling(Q,K,M_{spatial})\)
            \State \(\widehat{R} \gets Recall(\widehat{Q}, \widehat{K}, \widehat{M}_{temporal})\)
            \State \(\widetilde{R} \gets Recall(\widetilde{Q}, \widetilde{K}, \widetilde{M}_{spatial})\)
            \If {\((\widehat{R} \ge \alpha)\)}
                \State \(H \gets\) Temporal Head  \#\ \textit{high priority for Temporal}
            \ElsIf {\((\widetilde{R} \ge \alpha)\)}
                \State \(H \gets\) Spatial Head
            \Else
                \State \(H \gets\) Textural Head  
            \EndIf
            \State \Return \(H\)
        \end{algorithmic}
    \end{algorithm}
\end{minipage}\\
\\

\noindent (32, 30, 40) and (24, 30, 40) after VAE downsampling and patch embedding, respectively. For CogVideoX-5B, 45 frames are generated at $480 \times 720$, corresponding to a latent shape of (12, 30, 45).

\paragraph{Datasets and Benchmarks}
VBench \cite{huang2023vbenchcomprehensivebenchmarksuite} is a comprehensive benchmark suite for video generation tasks, systematically decomposing generation quality into 16 distinct evaluation dimensions. It further computes three weighted aggregated scores derived from these dimensions to holistically assess model performance. It consists of 946 prompts for all dimension evaluation. Video generation is a computation-heavy tasks, generating a four second 480p video costs for about 3 minutes. So we only use one random seed instead of five in all the following experiments. In the ablation study, for the sake of accelerating the experiments, we utilize 48 Sora prompts \cite{SoraPrompt2023}. And we use all VBench 946 prompts when comparing with other methods in section \ref{cmp_with_sota}.

\paragraph{Baselines}
We show the effectiveness and efficiency of RainFusion to compare it with other sparse or cache-based methods, including DiTFastAttn \cite{yuan2024ditfastattnattentioncompressiondiffusion}, $\Delta$-DiT\cite{chen2024deltadittrainingfreeaccelerationmethod}. For DiTFastAttn, we use their official configurations. For $\Delta$-DiT, we use the similar accelerate rate of RainFusion. For RainFusion, we set the sparsity to 50\% and we keep the first 10\% timesteps using dense calculation, which corresponds to about 1.85\(\times\) speedup in attention. Specifically, we set bandwidth = $\frac{1}{4}$ in both local and global pattern, corresponding to $\frac{9}{16}$ computation reduction. As for textural pattern, we reduce the key value tokens by half using the checkerboard layout as in Section \ref{head classification}.

\begin{figure*}[!]
  \centering
  \includegraphics[width=1.0\linewidth]{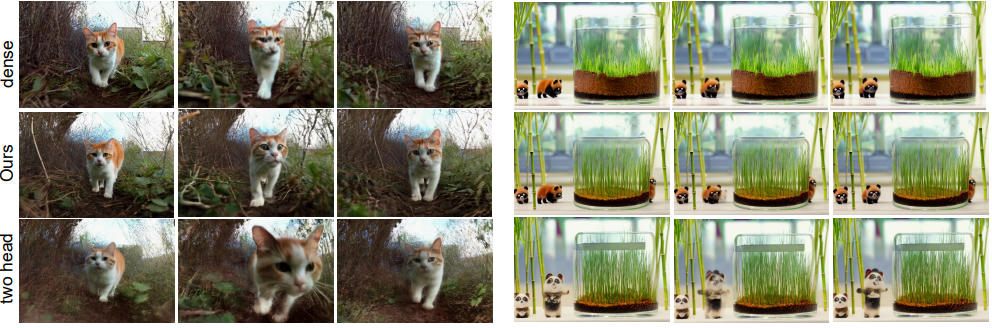}

   \caption{RainFusion video comparisons on CogVideoX-5B. Two Head means only use spatial and temporal head similar to SVG. We can see that RainFusion performs better than SVG and similar to baseline with 1.85x speedup. Left prompt: ``Animated scene features a close-up of a short fluffy monster kneeling beside a melting red candle". Right prompt: ``A petri dish with a bamboo forest growing within it that has tiny red pandas running around.”. }
   \label{fig:demo_video}
\end{figure*}

\begin{table*}
  \centering
  \resizebox{0.9\textwidth} {!}{
      \begin{tabular}{lcccccc}
      \toprule
        \multirow{2}{*}{Method} & \multirow{2}{*}{\begin{tabular}[c]{@{}c@{}}Average \\ Loss\end{tabular}} & \multirow{2}{*}{\begin{tabular}[c]{@{}c@{}}Subject\\ Consistency$\uparrow$\end{tabular}} & \multirow{2}{*}{\begin{tabular}[c]{@{}c@{}}Background\\ Consistency$\uparrow$\end{tabular}} & \multirow{2}{*}{\begin{tabular}[c]{@{}c@{}}Motion\\ Smoothness$\uparrow$\end{tabular}} & \multirow{2}{*}{\begin{tabular}[c]{@{}c@{}}Aesthetic\\ Quality$\uparrow$\end{tabular}} & \multirow{2}{*}{\begin{tabular}[c]{@{}c@{}}Imaging\\ Quality$\uparrow$\end{tabular}} \\
        
                                &                                                                          &                                                                                &                                                                                   &                                                                              &                                                                              &                                                                            \\
        \midrule
        baseline                & /                                                                        & 93.48                                                                          & 95.24                                                                             & 97.19                                                                        & 58.12                                                                        & 64.79                                                                      \\
        1.85\(\times\) RainFusion        & -0.21                                                                    & 93.27                                                                          & 95.31                                                                             & 97.23                                                                        & 58.11                                                                        & 63.81                                                                      \\
        2.50\(\times\) RainFusion        & -0.56                                                                    & 92.80                                                                           & 95.08                                                                             & 97.14                                                                        & 57.56                                                                        & 63.40                                                                      \\
        3.00\(\times\) RainFusion        & -1.36                                                                    & 92.36                                                                          & 94.85                                                                             & 97.27                                                                        & 55.20                                                                        & 62.30\\                                                                     
        \bottomrule
        \end{tabular}
    }
  \caption{Parameter Sensitivity Experiment Results. }
  \label{tab:sensitivity}
\end{table*}

\begin{figure*}[!]
  \centering
  \includegraphics[width=1.0\linewidth]{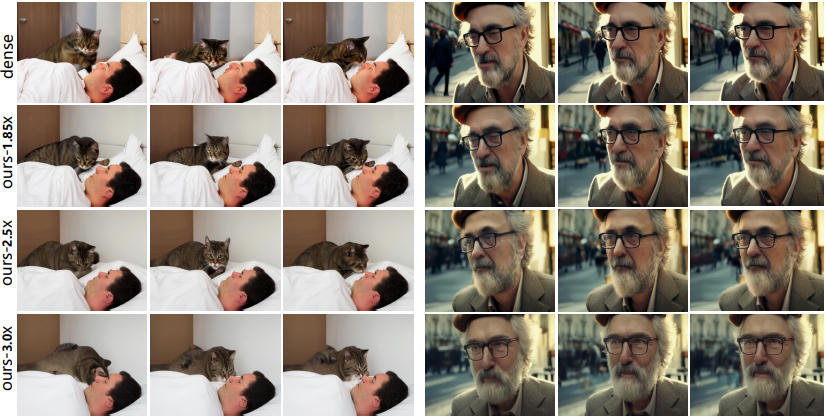}
   \caption{Video Comparison using CogVideoX-5B with different speedup ratio. Left prompt: ``A cat waking up its sleeping owner demanding breakfast. " Right Prompt: ``An extreme close-up of an gray-haired man with a beard in his 60s.".}
   \label{fig:Param_sensitivity}
\end{figure*}

\subsection{Ablation Study}
\paragraph{Component-wise Analysis} For RainFusion, there exists three kind of heads (Spatial, Temporal, Textural) as shown in Fig.\ref{fig:RainFusion} (c). We use ARM in Fig.\ref{fig:RainFusion} (b) to determine the pattern for each head. We do ablation study on the effectiveness of each head and how to estimate the patterns. 
We observe that there exists local sparse pattern inner the global sparse pattern as shown in the second row of Fig.\ref{fig:sparse_pattern}.
So if a head is above recall rate for both local and global pattern, we use local pattern first to cover more active region. We use global sampling as described in \ref{method:arm} to get global pattern recall.
To estimate local pattern recall, we compare the global sampling method and local sampling method(considering only the first frame tokens). As shown in Tab.\ref{tab:ablation}, we can see that with all three mask and with local estimation get the best results. 

Specifically, for CogVideoX-5B, using only two mask (same as SVG\cite{xi2025sparse} which only uses spatial and temporal head), the loss is 1.05\%. When adding textural head, we can get the best VBench score with the average loss  0.18\%. But if we change local estimate to global estimate, the loss is 0.42\%. And as shown in Fig.\ref{fig:demo_video}, we can see that RainFusion using three masks outperform using only two mask similar to SVG\cite{xi2025sparse}. Videos using our method is preserve more details with better imaging quality.

For OpenSoraPlan-1.2, it shows similar results. Using only two mask method drops by 1.29\%, when adding textural head, the result even perform better than baseline method. But when using global sampling, the result is a little worse, 0.33\% compared to 1.03\% improvement with local sampling in VBench score. So we use all three mask and local sampling as our default RainFusion configuration in the following sections.

\paragraph{Parameter Sensitivity} We test different sparse ratio by using different bandwidth and stride in spatial-temporal head and textural head, respectively. We test different RainFusion configuration in CogVideoX-5B of speedup 2.5\(\times\) and 3.0\(\times\) by setting the bandwidth of spatial and temporal head to 0.18 and 0.13, and textural stride to 3 and 4, respectively. As shown in Tab.\ref{tab:sensitivity}, our default setting can achieve 1.85\(\times\) speedup with 0.21\% average loss, while for 2.5\(\times\) and 3.0\(\times\) speedup, the loss is 0.56\% and 1.35\%. The 2.5\(\times\) RainFusion performs better than only local and global mask shown in Tab.\ref{tab:ablation} first row of CogVideoX-5B part, similar to SVG(1.85\(\times\) speedup). As for 3.0\(\times\) speedup, the loss is 1.35\%. It shows that our method can achieve more speedup with a tradeoff of accuracy. We can conclude that our method shows robust performance for different speedup ratio as shown in Fig.\ref{fig:Param_sensitivity}.

\subsection{Comparison with Baselines}
\label{cmp_with_sota}
\paragraph{Quantitative Results}
We compare RainFusion with sparse method DiTFastAttn and cache-based method $\Delta$-DiT. The result is shown in Table \ref{tab:cmp_with_sota}. We can see that in similar speedup, RainFusion performs best among other acceleration methods. 
Specifically, for CogVideoX-5B, RainFusion only drops by 0.28\% in VBench total score, while DiTFastAttn and $\Delta$-DiT drop by 5.37\% and 5.36\% respectively. For OpenSoraPlan-v1.2 and HunyuanVideo, the results is similar that our RainFusion performs best with -0.32\% and -0.4\% total score loss, respectively.

\paragraph{Qualitative Analysis and Integrability}
 As shown in Fig.\ref{fig:cmp_with_sota}, RainFusion achieves the best visual quality among all methods while DiTFastAttn and $\Delta$-DiT suffer from noise patch or inconsistency subjects. Notably, RainFusion is orthogonal to other acceleration approaches like cached-based method and can be combined to achieve a multiplicative speedup as shown in Fig.\ref{fig:hy_720p}. 
 For example, integrating RainFusion-1.84\(\times\) with $\Delta$-DiT 1.3\(\times\) yields a total speedup of 2.4\(\times\) on HunyuanVideo. As detailed in Tab.\ref{tab:ablation}, RainFusion+ variant employs dynamic bandwidth selection (0.5, 0.25, 0.125) across different attention heads. We determine the optimal bandwidth for each head by selecting the minimum value that maintains 90\% recall. Combining RainFusion+ with $\Delta$-DiT results in a minor -0.49\% performance drop, demonstrating its practical feasibility.

\section{Conclusion}
\label{sec:cls}
We introduce RainFusion, which utilizes spatial, temporal and textural sparsity in video generation models. Experiments demonstrate that RainFusion can achieve significant speed up in several video generation models with negligible quality loss (\textasciitilde\,0.2\% loss on VBench score). Our method is training-free and calibration-free, making it a plug-and-play tools to speed up video generation models. For future work, we will dive deeper to improve the sparsity ratio while preserving video quality and try to improve the video quality with fine-tuning.
{
    \small
    \bibliographystyle{ieeenat_fullname}
    \bibliography{main}

\begin{thebibliography}{42}
\providecommand{\natexlab}[1]{#1}
\providecommand{\url}[1]{\texttt{#1}}
\expandafter\ifx\csname urlstyle\endcsname\relax
  \providecommand{\doi}[1]{doi: #1}\else
  \providecommand{\doi}{doi: \begingroup \urlstyle{rm}\Url}\fi

\bibitem[Blattmann et~al.(2023{\natexlab{a}})Blattmann, Dockhorn, Kulal,
  Mendelevitch, Kilian, Lorenz, Levi, English, Voleti, Letts, Jampani, and
  Rombach]{blattmann2023stablevideodiffusionscaling}
Andreas Blattmann, Tim Dockhorn, Sumith Kulal, Daniel Mendelevitch, Maciej
  Kilian, Dominik Lorenz, Yam Levi, Zion English, Vikram Voleti, Adam Letts,
  Varun Jampani, and Robin Rombach.
\newblock Stable video diffusion: Scaling latent video diffusion models to
  large datasets, 2023{\natexlab{a}}.

\bibitem[Blattmann et~al.(2023{\natexlab{b}})Blattmann, Dockhorn, Kulal,
  Mendelevitch, Kilian, Lorenz, Levi, English, Voleti, Letts,
  et~al.]{blattmann2023stable}
Andreas Blattmann, Tim Dockhorn, Sumith Kulal, Daniel Mendelevitch, Maciej
  Kilian, Dominik Lorenz, Yam Levi, Zion English, Vikram Voleti, Adam Letts,
  et~al.
\newblock Stable video diffusion: Scaling latent video diffusion models to
  large datasets.
\newblock \emph{arXiv preprint arXiv:2311.15127}, 2023{\natexlab{b}}.

\bibitem[Bolya and Hoffman(2023)]{bolya2023token}
Daniel Bolya and Judy Hoffman.
\newblock Token merging for fast stable diffusion.
\newblock In \emph{Proceedings of the IEEE/CVF conference on computer vision
  and pattern recognition}, pages 4599--4603, 2023.

\bibitem[Chen et~al.(2024{\natexlab{a}})Chen, Ge, Xie, Wu, Yao, Ren, Wang, Luo,
  Lu, and Li]{chen2024pixart}
Junsong Chen, Chongjian Ge, Enze Xie, Yue Wu, Lewei Yao, Xiaozhe Ren, Zhongdao
  Wang, Ping Luo, Huchuan Lu, and Zhenguo Li.
\newblock Pixart-$\sigma$: Weak-to-strong training of diffusion transformer for
  4k text-to-image generation.
\newblock In \emph{European Conference on Computer Vision}, pages 74--91.
  Springer, 2024{\natexlab{a}}.

\bibitem[Chen et~al.(2024{\natexlab{b}})Chen, Shen, Ye, Cao, Tu, Bouganis,
  Zhao, and Chen]{chen2024delta}
Pengtao Chen, Mingzhu Shen, Peng Ye, Jianjian Cao, Chongjun Tu, Christos-Savvas
  Bouganis, Yiren Zhao, and Tao Chen.
\newblock $\delta $-dit: A training-free acceleration method tailored for
  diffusion transformers.
\newblock \emph{arXiv preprint arXiv:2406.01125}, 2024{\natexlab{b}}.

\bibitem[Chen et~al.(2024{\natexlab{c}})Chen, Shen, Ye, Cao, Tu, Bouganis,
  Zhao, and Chen]{chen2024deltadittrainingfreeaccelerationmethod}
Pengtao Chen, Mingzhu Shen, Peng Ye, Jianjian Cao, Chongjun Tu, Christos-Savvas
  Bouganis, Yiren Zhao, and Tao Chen.
\newblock $\delta$-dit: A training-free acceleration method tailored for
  diffusion transformers, 2024{\natexlab{c}}.

\bibitem[Chen et~al.(2023)Chen, Liu, Tang, Yi, Zhao, and
  Han]{chen2023sparsevit}
Xuanyao Chen, Zhijian Liu, Haotian Tang, Li Yi, Hang Zhao, and Song Han.
\newblock Sparsevit: Revisiting activation sparsity for efficient
  high-resolution vision transformer.
\newblock In \emph{Proceedings of the IEEE/CVF Conference on Computer Vision
  and Pattern Recognition}, pages 2061--2070, 2023.

\bibitem[Dhariwal et~al.(2022)Dhariwal, Ho, Jain, and
  Abbeel]{dhariwal2022guided}
Prafulla Dhariwal, Jonathan Ho, Ajay Jain, and Pieter Abbeel.
\newblock Guided diffusion models.
\newblock In \emph{NeurIPS}, 2022.

\bibitem[Henschel et~al.(2024)Henschel, Khachatryan, Hayrapetyan, Poghosyan,
  Tadevosyan, Wang, Navasardyan, and
  Shi]{henschel2024streamingt2vconsistentdynamicextendable}
Roberto Henschel, Levon Khachatryan, Daniil Hayrapetyan, Hayk Poghosyan, Vahram
  Tadevosyan, Zhangyang Wang, Shant Navasardyan, and Humphrey Shi.
\newblock Streamingt2v: Consistent, dynamic, and extendable long video
  generation from text, 2024.

\bibitem[Ho et~al.(2020)Ho, Jain, and Abbeel]{ho2020denoising}
Jonathan Ho, Ajay Jain, and Pieter Abbeel.
\newblock Denoising diffusion probabilistic models.
\newblock \emph{Advances in neural information processing systems},
  33:\penalty0 6840--6851, 2020.

\bibitem[Huang et~al.(2023)Huang, He, Yu, Zhang, Si, Jiang, Zhang, Wu, Jin,
  Chanpaisit, Wang, Chen, Wang, Lin, Qiao, and
  Liu]{huang2023vbenchcomprehensivebenchmarksuite}
Ziqi Huang, Yinan He, Jiashuo Yu, Fan Zhang, Chenyang Si, Yuming Jiang, Yuanhan
  Zhang, Tianxing Wu, Qingyang Jin, Nattapol Chanpaisit, Yaohui Wang, Xinyuan
  Chen, Limin Wang, Dahua Lin, Yu Qiao, and Ziwei Liu.
\newblock Vbench: Comprehensive benchmark suite for video generative models,
  2023.

\bibitem[Jiang et~al.(2024)Jiang, Li, Zhang, Wu, Luo, Ahn, Han, Abdi, Li, Lin,
  Yang, and Qiu]{jiang2024minference10acceleratingprefilling}
Huiqiang Jiang, Yucheng Li, Chengruidong Zhang, Qianhui Wu, Xufang Luo, Surin
  Ahn, Zhenhua Han, Amir~H. Abdi, Dongsheng Li, Chin-Yew Lin, Yuqing Yang, and
  Lili Qiu.
\newblock Minference 1.0: Accelerating pre-filling for long-context llms via
  dynamic sparse attention, 2024.

\bibitem[Kahatapitiya et~al.(2024)Kahatapitiya, Liu, He, Liu, Jia, Zhang, Ryoo,
  and Xie]{kahatapitiya2024adaptivecachingfastervideo}
Kumara Kahatapitiya, Haozhe Liu, Sen He, Ding Liu, Menglin Jia, Chenyang Zhang,
  Michael~S. Ryoo, and Tian Xie.
\newblock Adaptive caching for faster video generation with diffusion
  transformers, 2024.

\bibitem[Kong et~al.(2025)Kong, Tian, Zhang, Min, Dai, Zhou, Xiong, Li, Wu,
  Zhang, Wu, Lin, Yuan, Long, Wang, Wang, Li, Huang, Yang, Tan, Wang, Song,
  Bai, Wu, Xue, Wang, Wang, Liu, Li, Li, Wang, Yu, Deng, Li, Chen, Cui, Peng,
  Yu, He, Xu, Zhou, Xu, Tao, Lu, Liu, Zhou, Wang, Yang, Wang, Liu, Jiang, and
  Zhong]{kong2025hunyuanvideosystematicframeworklarge}
Weijie Kong, Qi Tian, Zijian Zhang, Rox Min, Zuozhuo Dai, Jin Zhou, Jiangfeng
  Xiong, Xin Li, Bo Wu, Jianwei Zhang, Kathrina Wu, Qin Lin, Junkun Yuan,
  Yanxin Long, Aladdin Wang, Andong Wang, Changlin Li, Duojun Huang, Fang Yang,
  Hao Tan, Hongmei Wang, Jacob Song, Jiawang Bai, Jianbing Wu, Jinbao Xue, Joey
  Wang, Kai Wang, Mengyang Liu, Pengyu Li, Shuai Li, Weiyan Wang, Wenqing Yu,
  Xinchi Deng, Yang Li, Yi Chen, Yutao Cui, Yuanbo Peng, Zhentao Yu, Zhiyu He,
  Zhiyong Xu, Zixiang Zhou, Zunnan Xu, Yangyu Tao, Qinglin Lu, Songtao Liu, Dax
  Zhou, Hongfa Wang, Yong Yang, Di Wang, Yuhong Liu, Jie Jiang, and Caesar
  Zhong.
\newblock Hunyuanvideo: A systematic framework for large video generative
  models, 2025.

\bibitem[Li et~al.(2023)Li, Li, Zheng, Wu, Xiao, Wang, Zheng, Pan, Chao, and
  Ji]{li2023autodiffusiontrainingfreeoptimizationtime}
Lijiang Li, Huixia Li, Xiawu Zheng, Jie Wu, Xuefeng Xiao, Rui Wang, Min Zheng,
  Xin Pan, Fei Chao, and Rongrong Ji.
\newblock Autodiffusion: Training-free optimization of time steps and
  architectures for automated diffusion model acceleration, 2023.

\bibitem[Liu et~al.(2024{\natexlab{a}})Liu, Zhang, Wang, Wei, Qiu, Zhao, Zhang,
  Ye, and Wan]{liu2024timestepembeddingtellsits}
Feng Liu, Shiwei Zhang, Xiaofeng Wang, Yujie Wei, Haonan Qiu, Yuzhong Zhao,
  Yingya Zhang, Qixiang Ye, and Fang Wan.
\newblock Timestep embedding tells: It's time to cache for video diffusion
  model, 2024{\natexlab{a}}.

\bibitem[Liu et~al.(2024{\natexlab{b}})Liu, Zhang, Xie, Faccio, Xu, Xiang,
  Shou, Perez-Rua, and Schmidhuber]{liu2024faster}
Haozhe Liu, Wentian Zhang, Jinheng Xie, Francesco Faccio, Mengmeng Xu, Tao
  Xiang, Mike~Zheng Shou, Juan-Manuel Perez-Rua, and J{\"u}rgen Schmidhuber.
\newblock Faster diffusion via temporal attention decomposition.
\newblock \emph{arXiv e-prints}, pages arXiv--2404, 2024{\natexlab{b}}.

\bibitem[Ma et~al.(2025)Ma, Tong, Jia, Hu, Su, Zhang, Yang, Li, Jaakkola, Jia,
  et~al.]{ma2025inference}
Nanye Ma, Shangyuan Tong, Haolin Jia, Hexiang Hu, Yu-Chuan Su, Mingda Zhang,
  Xuan Yang, Yandong Li, Tommi Jaakkola, Xuhui Jia, et~al.
\newblock Inference-time scaling for diffusion models beyond scaling denoising
  steps.
\newblock \emph{arXiv preprint arXiv:2501.09732}, 2025.

\bibitem[Ma et~al.(2024{\natexlab{a}})Ma, Fang, Mi, and Wang]{ma2024learning}
Xinyin Ma, Gongfan Fang, Michael~Bi Mi, and Xinchao Wang.
\newblock Learning-to-cache: Accelerating diffusion transformer via layer
  caching.
\newblock \emph{arXiv preprint arXiv:2406.01733}, 2024{\natexlab{a}}.

\bibitem[Ma et~al.(2024{\natexlab{b}})Ma, Fang, and Wang]{ma2023deepcache}
Xinyin Ma, Gongfan Fang, and Xinchao Wang.
\newblock Deepcache: Accelerating diffusion models for free.
\newblock In \emph{The IEEE/CVF Conference on Computer Vision and Pattern
  Recognition}, 2024{\natexlab{b}}.

\bibitem[Ma et~al.(2024{\natexlab{c}})Ma, Wang, Jia, Chen, Liu, Li, Chen, and
  Qiao]{ma2024lattelatentdiffusiontransformer}
Xin Ma, Yaohui Wang, Gengyun Jia, Xinyuan Chen, Ziwei Liu, Yuan-Fang Li,
  Cunjian Chen, and Yu Qiao.
\newblock Latte: Latent diffusion transformer for video generation,
  2024{\natexlab{c}}.

\bibitem[Nichol and Dhariwal(2021)]{nichol2021improved}
Alexander~Quinn Nichol and Prafulla Dhariwal.
\newblock Improved denoising diffusion probabilistic models.
\newblock In \emph{International conference on machine learning}, pages
  8162--8171. PMLR, 2021.

\bibitem[Open-Sora(2024)]{SoraPrompt2023}
Open-Sora.
\newblock Sora prompt, 2024.

\bibitem[OpenAI(2024)]{OpenAISora2024}
OpenAI.
\newblock Openai. sora, 2024.

\bibitem[Peebles and Xie(2023)]{peebles2023scalable}
William Peebles and Saining Xie.
\newblock Scalable diffusion models with transformers.
\newblock In \emph{Proceedings of the IEEE/CVF International Conference on
  Computer Vision}, pages 4195--4205, 2023.

\bibitem[Plan(2024)]{OpenSoraPlan2024}
Open-Sora Plan.
\newblock Open-sora plan, 2024.

\bibitem[Podell et~al.(2023)Podell, English, Lacey, Blattmann, Dockhorn,
  M{\"u}ller, Penna, and Rombach]{podell2023sdxl}
Dustin Podell, Zion English, Kyle Lacey, Andreas Blattmann, Tim Dockhorn, Jonas
  M{\"u}ller, Joe Penna, and Robin Rombach.
\newblock Sdxl: Improving latent diffusion models for high-resolution image
  synthesis.
\newblock \emph{arXiv preprint arXiv:2307.01952}, 2023.

\bibitem[Polyak et~al.(2024)Polyak, Zohar, Brown, Tjandra, Sinha, Lee, Vyas,
  Shi, Ma, Chuang, Yan, Choudhary, Wang, Sethi, Pang, Ma, Misra, Hou, Wang,
  Jagadeesh, Li, Zhang, Singh, Williamson, Le, Yu, Singh, Zhang, Vajda, Duval,
  Girdhar, Sumbaly, Rambhatla, Tsai, Azadi, Datta, Chen, Bell, Ramaswamy,
  Sheynin, Bhattacharya, Motwani, Xu, Li, Hou, Hsu, Yin, Dai, Taigman, Luo,
  Liu, Wu, Zhao, Kirstain, He, He, Pumarola, Thabet, Sanakoyeu, Mallya, Guo,
  Araya, Kerr, Wood, Liu, Peng, Vengertsev, Schonfeld, Blanchard, Juefei-Xu,
  Nord, Liang, Hoffman, Kohler, Fire, Sivakumar, Chen, Yu, Gao, Georgopoulos,
  Moritz, Sampson, Li, Parmeggiani, Fine, Fowler, Petrovic, and
  Du]{polyak2024moviegencastmedia}
Adam Polyak, Amit Zohar, Andrew Brown, Andros Tjandra, Animesh Sinha, Ann Lee,
  Apoorv Vyas, Bowen Shi, Chih-Yao Ma, Ching-Yao Chuang, David Yan, Dhruv
  Choudhary, Dingkang Wang, Geet Sethi, Guan Pang, Haoyu Ma, Ishan Misra, Ji
  Hou, Jialiang Wang, Kiran Jagadeesh, Kunpeng Li, Luxin Zhang, Mannat Singh,
  Mary Williamson, Matt Le, Matthew Yu, Mitesh~Kumar Singh, Peizhao Zhang,
  Peter Vajda, Quentin Duval, Rohit Girdhar, Roshan Sumbaly, Sai~Saketh
  Rambhatla, Sam Tsai, Samaneh Azadi, Samyak Datta, Sanyuan Chen, Sean Bell,
  Sharadh Ramaswamy, Shelly Sheynin, Siddharth Bhattacharya, Simran Motwani,
  Tao Xu, Tianhe Li, Tingbo Hou, Wei-Ning Hsu, Xi Yin, Xiaoliang Dai, Yaniv
  Taigman, Yaqiao Luo, Yen-Cheng Liu, Yi-Chiao Wu, Yue Zhao, Yuval Kirstain,
  Zecheng He, Zijian He, Albert Pumarola, Ali Thabet, Artsiom Sanakoyeu, Arun
  Mallya, Baishan Guo, Boris Araya, Breena Kerr, Carleigh Wood, Ce Liu, Cen
  Peng, Dimitry Vengertsev, Edgar Schonfeld, Elliot Blanchard, Felix Juefei-Xu,
  Fraylie Nord, Jeff Liang, John Hoffman, Jonas Kohler, Kaolin Fire, Karthik
  Sivakumar, Lawrence Chen, Licheng Yu, Luya Gao, Markos Georgopoulos, Rashel
  Moritz, Sara~K. Sampson, Shikai Li, Simone Parmeggiani, Steve Fine, Tara
  Fowler, Vladan Petrovic, and Yuming Du.
\newblock Movie gen: A cast of media foundation models, 2024.

\bibitem[Rombach et~al.(2022)Rombach, Blattmann, Lorenz, Esser, and
  Ommer]{rombach2022high}
Robin Rombach, Andreas Blattmann, Dominik Lorenz, Patrick Esser, and Bj{\"o}rn
  Ommer.
\newblock High-resolution image synthesis with latent diffusion models.
\newblock In \emph{Proceedings of the IEEE/CVF conference on computer vision
  and pattern recognition}, pages 10684--10695, 2022.

\bibitem[Sabour et~al.(2024)Sabour, Fidler, and
  Kreis]{sabour2024alignstepsoptimizingsampling}
Amirmojtaba Sabour, Sanja Fidler, and Karsten Kreis.
\newblock Align your steps: Optimizing sampling schedules in diffusion models,
  2024.

\bibitem[Smith et~al.(2024)Smith, Saxena, and Saha]{smith2024todo}
Ethan Smith, Nayan Saxena, and Aninda Saha.
\newblock Todo: Token downsampling for efficient generation of high-resolution
  images.
\newblock \emph{arXiv preprint arXiv:2402.13573}, 2024.

\bibitem[Tang et~al.(2024)Tang, Lin, Lin, Han, Hong, Yao, and
  Wang]{tang2024razorattention}
Hanlin Tang, Yang Lin, Jing Lin, Qingsen Han, Shikuan Hong, Yiwu Yao, and
  Gongyi Wang.
\newblock Razorattention: Efficient kv cache compression through retrieval
  heads.
\newblock \emph{arXiv preprint arXiv:2407.15891}, 2024.

\bibitem[Tian et~al.(2024)Tian, Tu, Chen, Hu, Xu, and Wang]{tian2024u}
Yuchuan Tian, Zhijun Tu, Hanting Chen, Jie Hu, Chao Xu, and Yunhe Wang.
\newblock U-dits: Downsample tokens in u-shaped diffusion transformers.
\newblock \emph{arXiv preprint arXiv:2405.02730}, 2024.

\bibitem[Tran et~al.(2025)Tran, MH~Nguyen, Nguyen, Nguyen, Le, Xie, Sonntag,
  Zou, Nguyen, and Niepert]{tran2025accelerating}
Chau Tran, Duy MH~Nguyen, Manh-Duy Nguyen, TrungTin Nguyen, Ngan Le, Pengtao
  Xie, Daniel Sonntag, James~Y Zou, Binh Nguyen, and Mathias Niepert.
\newblock Accelerating transformers with spectrum-preserving token merging.
\newblock \emph{Advances in Neural Information Processing Systems},
  37:\penalty0 30772--30810, 2025.

\bibitem[Wang et~al.(2024)Wang, Liu, Kang, Li, Lin, Jha, and
  Liu]{wang2024attention}
Hongjie Wang, Difan Liu, Yan Kang, Yijun Li, Zhe Lin, Niraj~K Jha, and Yuchen
  Liu.
\newblock Attention-driven training-free efficiency enhancement of diffusion
  models.
\newblock In \emph{Proceedings of the IEEE/CVF Conference on Computer Vision
  and Pattern Recognition}, pages 16080--16089, 2024.

\bibitem[Wimbauer et~al.(2024)Wimbauer, Wu, Schoenfeld, Dai, Hou, He,
  Sanakoyeu, Zhang, Tsai, Kohler, et~al.]{wimbauer2024cache}
Felix Wimbauer, Bichen Wu, Edgar Schoenfeld, Xiaoliang Dai, Ji Hou, Zijian He,
  Artsiom Sanakoyeu, Peizhao Zhang, Sam Tsai, Jonas Kohler, et~al.
\newblock Cache me if you can: Accelerating diffusion models through block
  caching.
\newblock In \emph{Proceedings of the IEEE/CVF Conference on Computer Vision
  and Pattern Recognition}, pages 6211--6220, 2024.

\bibitem[Wu et~al.(2024)Wu, Xu, Le, and Samaras]{wu2024importance}
Haoyu Wu, Jingyi Xu, Hieu Le, and Dimitris Samaras.
\newblock Importance-based token merging for diffusion models.
\newblock \emph{arXiv preprint arXiv:2411.16720}, 2024.

\bibitem[Xi et~al.(2025)Xi, Yang, Zhao, Xu, Li, Li, Lin, Cai, Zhang, Li,
  et~al.]{xi2025sparse}
Haocheng Xi, Shuo Yang, Yilong Zhao, Chenfeng Xu, Muyang Li, Xiuyu Li, Yujun
  Lin, Han Cai, Jintao Zhang, Dacheng Li, et~al.
\newblock Sparse videogen: Accelerating video diffusion transformers with
  spatial-temporal sparsity.
\newblock \emph{arXiv preprint arXiv:2502.01776}, 2025.

\bibitem[Yang et~al.(2024)Yang, Teng, Zheng, Ding, Huang, Xu, Yang, Hong,
  Zhang, Feng, Yin, Gu, Zhang, Wang, Cheng, Liu, Xu, Dong, and
  Tang]{yang2024cogvideoxtexttovideodiffusionmodels}
Zhuoyi Yang, Jiayan Teng, Wendi Zheng, Ming Ding, Shiyu Huang, Jiazheng Xu,
  Yuanming Yang, Wenyi Hong, Xiaohan Zhang, Guanyu Feng, Da Yin, Xiaotao Gu,
  Yuxuan Zhang, Weihan Wang, Yean Cheng, Ting Liu, Bin Xu, Yuxiao Dong, and Jie
  Tang.
\newblock Cogvideox: Text-to-video diffusion models with an expert transformer,
  2024.

\bibitem[Yuan et~al.(2024)Yuan, Zhang, Lu, Ning, Zhang, Zhao, Yan, Dai, and
  Wang]{yuan2024ditfastattnattentioncompressiondiffusion}
Zhihang Yuan, Hanling Zhang, Pu Lu, Xuefei Ning, Linfeng Zhang, Tianchen Zhao,
  Shengen Yan, Guohao Dai, and Yu Wang.
\newblock Ditfastattn: Attention compression for diffusion transformer models,
  2024.

\bibitem[Zhang and Papyan(2024)]{zhang2024oats}
Stephen Zhang and Vardan Papyan.
\newblock Oats: Outlier-aware pruning through sparse and low rank
  decomposition.
\newblock \emph{arXiv preprint arXiv:2409.13652}, 2024.

\bibitem[Zheng et~al.(2024)Zheng, Peng, Yang, Shen, Li, Liu, Zhou, Li, and
  You]{zheng2024open}
Zangwei Zheng, Xiangyu Peng, Tianji Yang, Chenhui Shen, Shenggui Li, Hongxin
  Liu, Yukun Zhou, Tianyi Li, and Yang You.
\newblock Open-sora: Democratizing efficient video production for all.
\newblock \emph{arXiv preprint arXiv:2412.20404}, 2024.

\end{thebibliography}
}

\end{document}